\documentclass[runningheads]{llncs}

\usepackage[T1]{fontenc}
\usepackage{graphicx}
\usepackage{amsmath}
\usepackage{amssymb}
\usepackage{hyperref}
\hypersetup{
  colorlinks=false,
  citebordercolor={0 1 0},
  linkbordercolor={1 0 0},
  urlbordercolor={0 0 1},
  pdfborder={0 0 1}
}

\begin{document}

\title{MORPHA: Morphology-Constrained Training and the Limits of Cross-Acquisition Transfer in Low-Resource Malaria Microscopy}
\titlerunning{MORPHA: Limits of Cross-Acquisition Transfer in Malaria Microscopy}

\author{%
Favour Okechukwu Igwezeke\inst{1}\orcidID{0009-0001-9907-4696} \and
Chikodili Helen Ugwuishiwu\inst{2}\orcidID{0000-0003-3166-6633} \and
Joseph Uzochukwu Emesiani\inst{2}\orcidID{0009-0003-4220-9195} \and
Samuel Ifebuche Agada\inst{3}\orcidID{0009-0005-4043-4174} \and
Ekenechukwu Lilian Anozie\inst{4}\orcidID{0009-0006-8430-7908} \and
Mary Ofuru Kama\inst{5}\orcidID{0009-0009-0979-2220} \and
Adaobi Chiazor Emegoakor\inst{6}\orcidID{0009-0008-2874-6708}}
\authorrunning{F. O. Igwezeke et al.}
\institute{%
Faculty of Pharmaceutical Sciences, University of Nigeria, Nsukka, Nigeria\\ \email{favour.igwezeke.249461@unn.edu.ng} \and
Department of Computer Science, University of Nigeria, Nsukka, Nigeria \and
Department of Radiology, National Hospital Abuja, Nigeria \and
Department of Computer Science, State University of Medical and Applied Sciences, Igbo-Eno, Enugu State, Nigeria \and
Department of Software Engineering, Veritas University, Abuja, Nigeria \and
Department of Radiology, Nnamdi Azikiwe University Teaching Hospital, Nnewi, Anambra State, Nigeria}

\maketitle

\begin{abstract}
In low-resource malaria microscopy, a model trained on one smear preparation routinely meets images from another, and how well morphology-based constraints transfer across this acquisition gap is unclear. We study this on real African field microscopy from Uganda (Lacuna), asking where encoding measured parasite morphology as a training constraint improves cross-acquisition transfer and where generic regularisation suffices. We present \textbf{MORPHA}, a morphological consistency constraint that derives stage-conditional statistics from the stage-annotated BBBC041 dataset and penalises predictions that deviate from them. Defined uniformly across binary, object-level, and stage-aware regimes without changing architecture or inference, it shapes training in the binary regime. The detection regime is a mapped boundary. On transfer from thin-smear cells to thick-smear field images, the constraint reduces the binary-classification generalisation drop by 30.8\% (F1 0.578 to 0.699) at negligible within-domain cost and lowers in-distribution calibration error by 49\% (ECE 0.0162 to 0.0082). A content-free control applying the identical constraint to random statistics recovers less of the drop (25.3\% vs 30.8\%), indicating the measured content, not constraining alone, contributes to the gain. Two standard confidence regularisers exceed the constraint on raw transfer, locating where morphology adds value and where generic regularisation suffices. We map two deployment-relevant boundaries: thin-smear statistics do not transfer to thick-smear detection (trophozoite AP@0.50 falls to 0.000), and cross-acquisition pseudo-labelling fails before filtering applies. Together these yield a morphology-grounded consistency signal and evidence-based guidance for malaria dataset and model design in low-resource settings.

\end{abstract}

\section{Introduction}
Malaria diagnosis in many endemic regions still rests on light microscopy of Giemsa-stained blood smears. A single examination can confirm infection, identify the species, quantify parasite density, and characterise developmental stage~\cite{makanjuola2020}. That reach depends on skilled microscopists and sustained quality assurance~\cite{ngasala2019}. Reader agreement is weakest at the low densities where accurate counts matter most~\cite{omeara2005}. Where expertise is scarce, this variability erodes consistency and limits scale.

Deep learning eases some of these constraints, and recent systems approach expert microscopists on curated smears~\cite{kuo2020}. Their accuracy, however, is inherited from the annotations they train on, and those differ widely in richness. Binary labels record only whether a cell is infected, object-level boxes add localisation, and stage-aware boxes further distinguish rings, trophozoites, schizonts, and gametocytes~\cite{nakasi2024,soracardenas2025}. Stage-aware labels demand specialised expertise and are costlier to produce at scale, yet they carry prognostic information, such as mature schizonts, that binary labels cannot convey~\cite{vanwolfswinkel2012}. In most malaria-detection systems morphology enters only implicitly, as a feature the network may or may not learn~\cite{poostchi2018,abdurahman2021,fuhad2020,nakasi2020,boit2024}. Morphology-guided supervision aids clinical interpretability~\cite{yousaf2026}, but whether it transfers across smear preparations is rarely tested head-on.

We propose \textbf{MORPHA}, a morphological consistency constraint that encodes stage statistics from BBBC041~\cite{ljosa2012} as a soft training penalty, defined uniformly across all three annotation regimes with no change to architecture or inference. In the binary regime it acts as a prediction-level penalty that shapes training directly. In the detection regime it operates over fixed annotation boxes, treated as a mapped boundary. We use \textbf{MORPHA} to characterise, on real African field microscopy, where a morphological constraint improves cross-acquisition transfer and where generic regularisation suffices, rather than to claim a best-performing method. We evaluate transfer from thin-smear to thick-smear field images, characterise uncertainty with Monte Carlo Dropout~\cite{gal2016}, and use a content-free control plus a benchmark against standard regularisers.

We make three contributions. (1)~Using a content-free control, we show the gains come from the measured morphological content rather than the act of constraining, and we map where the constraint helps and where it does not. (2)~The constraint reduces the binary cross-dataset generalisation drop by 30.8\% at negligible within-domain cost and improves in-distribution calibration by 49\%. On raw transfer, two standard regularisers exceed it. (3)~We establish two deployment-relevant transfer boundaries: thin-smear statistics do not transfer to thick-smear detection, and cross-acquisition pseudo-labelling fails before morphological filtering can apply, together showing that thin-smear benchmarks do not proxy thick-smear field performance.

\section{Related Work}
Most malaria deep-learning systems treat morphology as an emergent feature rather than a supervised signal. Binary parasitized-versus-uninfected classifiers dominate: Fuhad et al.~\cite{fuhad2020} report 99.23\% on the NIH thin-smear set, and Boit \& Patil~\cite{boit2024} reach 97.68\% with their EDRI hybrid on the same data. Stage-aware supervision is rarer and harder. Chaudhry et al.~\cite{chaudhry2024} classify parasite type and life-cycle stage with a lightweight network across four datasets, while Abbas \& Dijkstra~\cite{abbas2020} obtain 82.7\% on three merged stages using random-forest classifiers, noting that sub-stages are mostly confused with neighbours. Closest to our framing, Yousaf et al.~\cite{yousaf2026} (MorphXAI) inject morphological supervision into a detection decoder but evaluate within a single preparation.

A recurring observation is that benchmark performance does not survive acquisition shift. Pongpanitanont et al.~\cite{pongpanitanont2026} emphasise that image-level splits inflate metrics and call for external validation across staining and sites. Poostchi et al.~\cite{poostchi2018} note that almost all deep-learning work targets thin smears, with thick-smear detection comparatively unexplored. Critically, nearly every study trains and tests within one smear preparation, leaving open whether morphological supervision transfers from thin to thick smears.

\section{Datasets}
\begin{figure}[htbp]
\centering
\includegraphics[width=\textwidth]{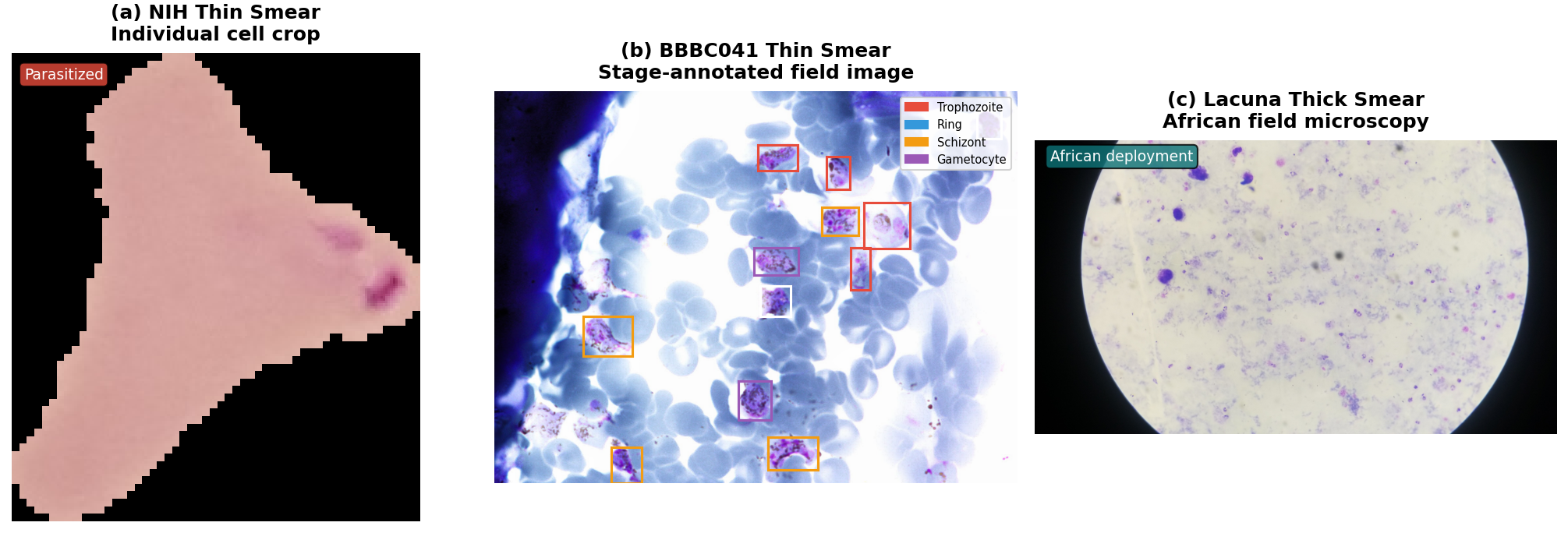}
\caption{Smear preparation creates a visual domain shift across the three datasets. (a)~NIH thin-smear individual cell crop with binary labels. (b)~BBBC041 thin-smear field image with stage-annotated boxes. (c)~Lacuna thick-smear African field microscopy, the primary transfer target.}
\label{fig:datasets}
\end{figure}

We use three public datasets, one per annotation regime (Fig.~\ref{fig:datasets}). \textbf{NIH Malaria Cell Images}~\cite{rajaraman2018} provides 27,558 segmented thin-smear cells, balanced 13,779/13,779 parasitized/uninfected, with image-level binary labels only (Fig.~\ref{fig:datasets}a). \textbf{BBBC041}~\cite{ljosa2012} provides stage-aware boxes and is the source of our morphological statistics (Fig.~\ref{fig:datasets}b). Following the dataset's training split we use 1,208 of its 1,328 images, divided 1,026/182 for training and validation. \textbf{Lacuna Malaria Blood Smear}~\cite{nakasi2024} provides 2,747 full-field thick- and thin-smear microscopy images from Uganda, captured by smartphone over a microscope eyepiece (Fig.~\ref{fig:datasets}c). Predominantly thick-smear, its field provenance makes it our primary transfer target.

\section{Methods}
\begin{figure}[htbp]
\centering
\includegraphics[width=\textwidth]{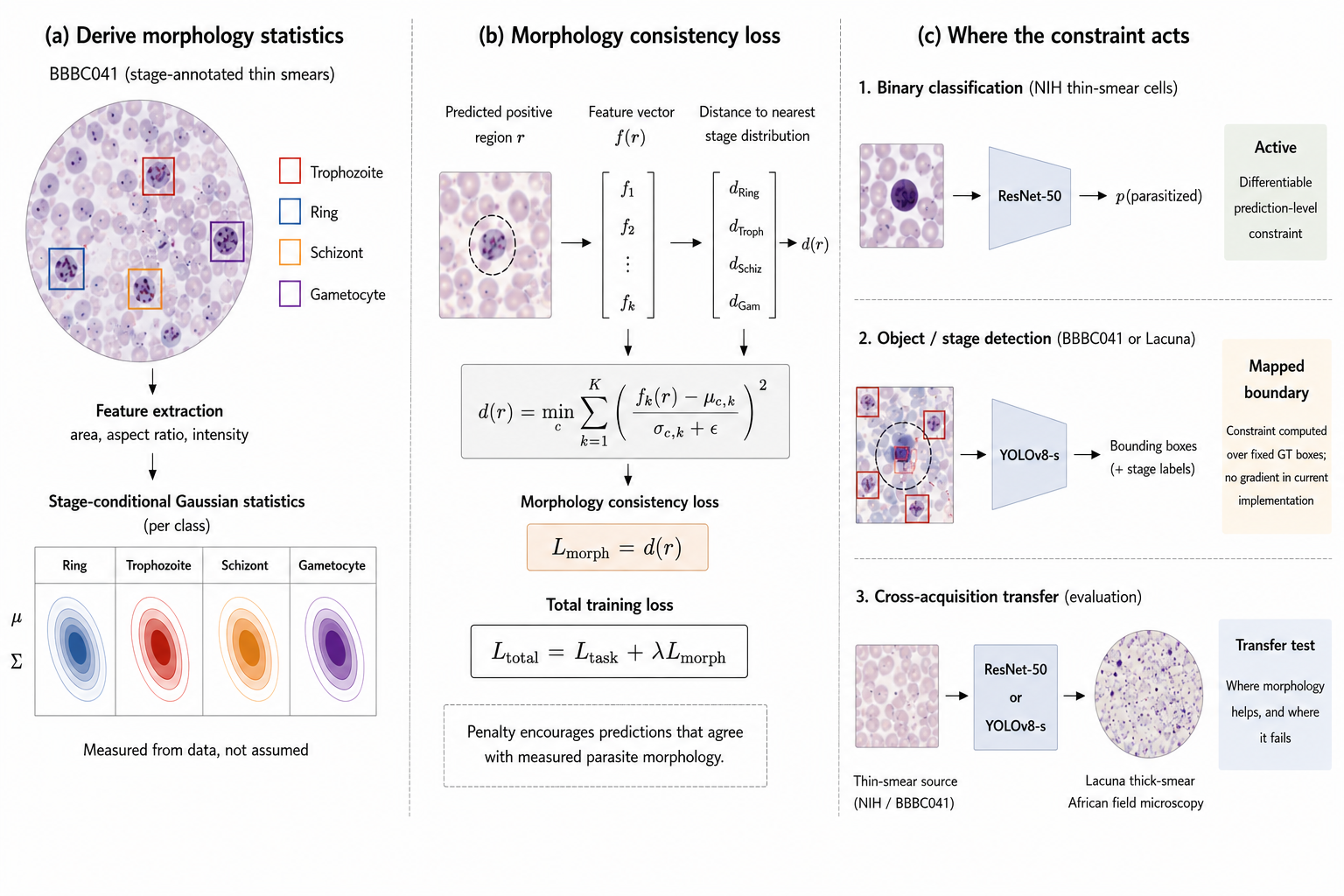}
\caption{Overview of MORPHA. (a)~Stage-conditional morphology statistics (area, aspect ratio, intensity) are derived from BBBC041 annotations and stored as per-class Gaussians. (b)~For a predicted positive region $r$, a Mahalanobis consistency score $d(r)$ to the nearest stage distribution is added to the task loss as a soft penalty, without changing architecture or inference. (c)~The constraint acts directly in the binary regime, is a mapped boundary in detection (fixed ground-truth boxes, no gradient), and is evaluated under cross-acquisition transfer from thin-smear sources to Lacuna African field microscopy.}
\label{fig:framework}
\end{figure}

\subsection{Morphological statistics}
We derive stage-conditional statistics from the BBBC041 training set (Fig.~\ref{fig:framework}a). For each of 79,672 annotated instances across six classes (446 difficult instances excluded), we extract five features: bounding-box area (px), aspect ratio, mean and standard deviation of normalised intensity, and area relative to the mean uninfected-RBC box (11,678~px\textsuperscript{2}). Intensity is the RGB mean divided by 255. Per-class means and standard deviations are stored as class-conditional Gaussian parameters (Table~\ref{tab:stats}). We derive these from data, not published descriptions, so they reflect the source imaging conditions, which ties them to a single preparation type.

\begin{table}[htbp]
\centering
\small
\caption{Morphological statistics from BBBC041. Rel.\ RBC = stage area / uninfected-RBC area. Mean $\pm$ SD.}
\label{tab:stats}
\begin{tabular}{lrrrrr}
\hline
Stage & N & BB Area px\textsuperscript{2} & Asp.\ Ratio & Rel.\ RBC & Intensity \\
\hline
Uninfected RBC & 77,420 & 11,678 $\pm$ 2,610 & 1.019 $\pm$ 0.157 & 1.000 & 0.529 $\pm$ 0.129 \\
Ring & 353 & 15,239 $\pm$ 3,064 & 1.006 $\pm$ 0.176 & 1.305 & 0.509 $\pm$ 0.087 \\
Trophozoite & 1,473 & 17,919 $\pm$ 4,746 & 1.026 $\pm$ 0.250 & 1.534 & 0.515 $\pm$ 0.100 \\
Schizont & 179 & 20,377 $\pm$ 4,562 & 1.009 $\pm$ 0.237 & 1.745 & 0.593 $\pm$ 0.102 \\
Gametocyte & 144 & 17,270 $\pm$ 3,316 & 1.027 $\pm$ 0.228 & 1.479 & 0.513 $\pm$ 0.083 \\
Leukocyte & 103 & 16,549 $\pm$ 8,092 & 1.029 $\pm$ 0.170 & 1.417 & 0.350 $\pm$ 0.106 \\
\hline
\end{tabular}
\end{table}

\subsection{Morphological consistency loss}
The morphological consistency score $d(r)$ is the minimum squared Mahalanobis distance from a predicted positive region $r$ to any parasite-stage entry, in the feature subspace available for the regime (Fig.~\ref{fig:framework}b):
\begin{equation}
d(r) = \min_{c} \sum_{k} \left( \frac{f_k(r) - \mu_{c,k}}{\sigma_{c,k} + \epsilon} \right)^{2}
\end{equation}
with $C = \{\text{ring, trophozoite, schizont, gametocyte}\}$ and $\epsilon = 10^{-6}$. For NIH, each image is a single cell with no box, so $d(r)$ uses image-level intensity only, weighted by the parasitized probability $p_1(r)$:
\begin{equation}
\mathcal{L}_{\mathrm{morph}}^{\mathrm{bin}} = \frac{1}{|B|} \sum_{r} p_1(r)\, d(r), \qquad d(r) = \min_{c} \left( \frac{I(r) - \mu_{c,I}}{\sigma_{c,I} + \epsilon} \right)^{2}
\end{equation}
For detection, $d(r)$ uses aspect ratio and intensity (box area is excluded, as coordinates are normalised to $640\times640$). As implemented, the detection term is computed over ground-truth boxes, fixed tensors that supply no gradient, as reported below. The objective is $\mathcal{L}_{\mathrm{total}} = \mathcal{L}_{\mathrm{task}} + \lambda \mathcal{L}_{\mathrm{morph}}$, with $\mathcal{L}_{\mathrm{task}}$ binary cross-entropy (NIH) or the YOLOv8 composite loss (detection). We set $\lambda = 0.1$ by grid search over $\{0.01, 0.1, 1.0\}$ on NIH validation, selecting on transfer to Lacuna rather than within-distribution performance.

\subsection{Architectures and uncertainty}
The binary model is ResNet-50~\cite{he2016}, ImageNet-pretrained at $224\times224$, with a two-class head, unmodified so that constrained-versus-unconstrained differences are attributable to the loss alone. The classification model has 23.5M parameters and runs single-image inference in 120~ms (single CPU thread, batch size 1, $n=100$), about 8 images per second without GPU acceleration. Detection uses the YOLOv8-small detector~\cite{jocher2023}, which has 11.1M parameters, with six classes for BBBC041 and two for Lacuna. For uncertainty, we apply Monte Carlo Dropout~\cite{gal2016}: we add dropout ($p=0.3$) before the ResNet-50 head and run $T=20$ stochastic passes. Uncertainty is the variance of the parasitized probability, and calibration is measured by Expected Calibration Error (ECE)~\cite{guo2017} over 15 bins. We also compute the Pearson correlation between $d(r)$ and uncertainty over predicted positives, interpreting its magnitude cautiously.

\subsection{Pseudo-labelling and training}
To test bridging at the data level, we apply morphology-filtered pseudo-labelling~\cite{lee2013}: the BBBC041 detector labels all 9,645 NIH parasitized cells. Predictions with a stage label, confidence above 0.15, and $d(r) \le \tau$ ($\tau = 0.237$, the 25th distance percentile) fine-tune a multi-task head added to the ResNet-50 backbone. All runs use PyTorch 2.0 on one Tesla T4 GPU, seed 42. Binary models train 50 epochs (AdamW, lr $1\times10^{-4}$, cosine, batch 32). Detection trains 100 epochs (SGD, lr 0.01, batch 8). NIH is split 70/15/15 stratified. Lacuna provides 2,432 matched images for classification evaluation and a separate 85/15 split for detection.

\section{Results}
\subsection{Within-dataset performance}
The constraint preserves within-domain accuracy. On NIH, Binary-MC reaches F1 $=0.9729$ against 0.9739 unconstrained, a difference of about four images in 4,136. Both detection models are identical with and without the constraint, because the detection term is computed over ground-truth boxes and supplies no gradient (Fig.~\ref{fig:framework}c). A prediction-level detection formulation is future work. Stage-aware detection reaches mAP50 $=0.806$, and within-Lacuna detection reaches 0.835.

\subsection{Cross-dataset generalisation}
The constraint substantially improves transfer to African field images (Table~\ref{tab:cross}). Binary-UC falls from F1 $=0.9739$ on NIH to 0.5783 on Lacuna. The constrained model holds within-domain F1 (0.9729) and reaches 0.6992 on Lacuna, cutting the generalisation drop from 0.3956 to 0.2737, a 30.8\% relative reduction. Two factors compound the gap: NIH cells are thin-smear crops while Lacuna images are full-field thick smears (Fig.~\ref{fig:datasets}a,~c), and the Table~\ref{tab:stats} statistics are specific to thin-smear staining. That a thin-smear-derived constraint still improves thick-smear transfer indicates a useful inductive bias, isolated by the content control below.

\begin{table}[htbp]
\centering
\small
\caption{Cross-dataset generalisation (thin-smear to Lacuna). Drop = within $-$ cross. $\ddagger$~BBBC041 trophozoite mAP50. $\dagger$~predicted, not measured.}
\label{tab:cross}
\begin{tabular}{lllrrr}
\hline
Model & Source & Target & Within & Cross & Drop \\
\hline
Binary-UC & NIH & Lacuna & 0.9739 & 0.5783 & 0.3956 \\
Binary-MC & NIH & Lacuna & 0.9729 & 0.6992 & 0.2737 \\
Stage-UC & BBBC041 & Lacuna & 0.832$\ddagger$ & 0.000 & 0.832 \\
Stage-MC$\dagger$ & BBBC041 & Lacuna & 0.832$\ddagger$ & 0.000$\dagger$ & 0.832$\dagger$ \\
\hline
\end{tabular}
\end{table}

The detection regime maps a transfer boundary rather than testing the constraint, since the detection term carries no gradient. Applied to Lacuna, the BBBC041 stage detector yields trophozoite AP@0.50 $=0.000$, with only 2 of 375 predictions reaching IoU $\ge 0.50$ against 14,409 ground-truth boxes, versus a within-Lacuna trophozoite mAP50 of 0.689. The boundary is asymmetric: binary classification loses roughly 40\% of F1 and the constraint recovers part of it, whereas detection loses all transferable capability.

\subsection{Uncertainty and calibration}
The constraint improves calibration: in-distribution ECE falls 49\% (0.0162 to 0.0082) and out-of-distribution ECE falls 27\% (0.4827 to 0.3517). OOD ECE stays high, with both models overconfident on Lacuna (mean confidence 0.947 UC, 0.906 MC) against far lower accuracy, which the constraint reduces but does not remove. The correlation between $d(r)$ and uncertainty is significant only for the constrained model ($r=0.047$, $p=0.033$ vs $r=0.028$, $p=0.215$) but small (about 0.2\% of variance), so we read it as a weak association. A more robust trend appears across consistency quintiles, where accuracy falls from 0.990 (most consistent) to 0.949 (least) and the $d(r)$--correctness correlation is $-0.097$ ($p<0.0001$) for the constrained model versus $-0.082$ for the unconstrained.

\subsection{Content and regulariser controls}
Two controls establish what the constraint contributes (Table~\ref{tab:controls}). Random statistics under the identical mechanism recover less of the drop (25.3\% vs 30.8\%) and lower cross-domain F1 (0.6813 vs 0.6992), so the measured content carries value beyond constraining. Label smoothing~\cite{szegedy2016} and a confidence penalty~\cite{pereyra2017}, neither using domain knowledge, exceed the constraint on raw transfer (drop reductions 46.4\% and 40.9\%) with lower OOD ECE. The constraint is thus not the strongest technique for raw generalisation here. Its distinct contribution is a consistency score in interpretable units of deviation from measured morphology, which the regularisers do not provide.

\begin{table}[htbp]
\centering
\small
\caption{Content-free (random) ablation and two regularisers. Lacuna matched set ($n=2{,}432$). Reduction = relative drop reduction vs Binary-UC. Single-pass ECE.}
\label{tab:controls}
\begin{tabular}{lrrrrr}
\hline
Model & NIH F1 & NIH ECE & Lac F1 & Lac ECE & Reduction \\
\hline
Binary-UC & 0.9739 & 0.0165 & 0.5783 & 0.4827 & --- \\
Binary-MC & 0.9729 & 0.0084 & 0.6992 & 0.3517 & 30.8\% \\
Binary-MC-Random & 0.9768 & 0.0071 & 0.6813 & 0.2974 & 25.3\% \\
Binary-LS (label smoothing) & 0.9692 & 0.0367 & 0.7570 & 0.2001 & 46.4\% \\
Binary-CP (confidence penalty) & 0.9741 & 0.0031 & 0.7404 & 0.2530 & 40.9\% \\
\hline
\end{tabular}
\end{table}

\subsection{Pseudo-labelling}
Cross-acquisition pseudo-labelling fails before filtering can act. The BBBC041 detector produces only 17 candidate stage predictions from 9,645 NIH cells (0.18\%), of which five survive filtering. The cause is a domain mismatch: the detector was trained on full-field smears of 50--200 cells, whereas NIH images are single cells filling the frame, so nearly all are labelled RBC. With five labels the stage head is effectively unsupervised, and the resulting model reaches F1 $=0.9719$ with a stage distribution uncorrelated with ground truth ($r=0.920$, $p=0.081$) and over-predicting schizonts (24.4\% vs 8.3\%). Bridging these modes requires explicit domain adaptation.

\subsection{Sensitivity to $\lambda$}
Sensitivity to the constraint weight shows three regimes. At $\lambda=0.01$ the constraint is too weak to steer training and transfer is worse than baseline (Lacuna F1 0.4868). At $\lambda=0.1$ the drop falls 30.8\% with negligible within-domain cost and a 49\% ECE improvement. At $\lambda=1.0$ the drop is lowest (0.2321) and ECE best (0.0067) but within-domain F1 falls by 0.010. We select $\lambda=0.1$ as the point where robustness improves without a meaningful within-domain cost. A stronger $\lambda=1.0$ suits settings that prioritise robustness over benchmark accuracy.

\section{Discussion}
The evidence supports a precise claim: the constraint improves binary cross-dataset transfer to African field microscopy by 30.8\% at negligible within-domain cost and improves calibration, and the content-free control shows the measured statistics, not constraining alone, drive the gain. Two standard regularisers exceed the constraint on raw transfer, so we position it not as a superior generalisation method but as one that adds an interpretable consistency signal, computed in units of deviation from measured morphology, that those regularisers do not provide. Whether that interpretability adds deployment value beyond calibration is not established here.

The smear-preparation boundary recurs across regimes: thin-smear-derived statistics and models transfer partially in the binary regime and not at all in detection, where a thin-smear detector essentially does not fire on thick smears. This is an empirical limit with direct deployment relevance, since thick smears are common in routine field screening. The main limitations are single runs per condition (directionally stable), $\lambda$ tuned for the constraint and not per baseline, the detection term carrying no gradient as implemented, proxy features used throughout, a weak morphology--uncertainty correlation, and a single backbone and thin-smear source per regime.

\section{Impact in Resource-Constrained Settings}
Stage-aware annotation requires expertise that is scarce in target settings, while binary labels are cheap. Incorporating measured stage morphology recovers part of the robustness that richer annotation would provide, though less than the two zero-domain-knowledge regularisers achieve. Under budget limits, label smoothing or a confidence penalty is therefore a sensible first step, with the constraint as a complementary, interpretable addition. Requiring no GPU at inference (120~ms per image on a single CPU thread), the classification model suits microscopy workstations without dedicated accelerators. Across every variant, models stayed overconfident on Lacuna, predicting confidently while far less accurate. Where confirmatory review is unavailable, such confident errors are a particular risk, regardless of the mitigation chosen. Finally, the failure of thin-smear detection on thick-smear images shows that thin-smear benchmarks cannot proxy deployment performance, arguing for deliberate inclusion of field-acquired imagery in future datasets.

\begin{credits}
\subsubsection{\ackname}
This work was carried out as part of the Sprint AI Training for African Medical Imaging Knowledge Translation (SPARK) Academy 2026 summer school on deep learning in medical imaging. The authors gratefully acknowledge the guidance and coordination of the SPARK Academy program leadership (Udunna C. Anazodo, Dong Zhang, Confidence Raymond, and Aondona M. Iorumbur), whose oversight of the program, mentorship, and securing of resources made this work possible. The authors also thank the SPARK Academy 2025 instructors for the background knowledge that informed this research, and Linshan Liu for administrative support. The authors acknowledge computational-infrastructure support from the Digital Research Alliance of Canada (the Alliance), the University of Washington Azure GenAI for Science Hub, and knowledge-translation support from the McGill University Doctoral Internship Program. Finally, the authors thank the Lacuna Fund for Health and Equity (PI: Udunna Anazodo, 0508-S-001), the Radiological Society of North America (RSNA) Research \& Education (R\&E) Foundation, and the National Science and Engineering Research Council of Canada (NSERC) Discovery Launch Supplement (PI: Udunna Anazodo, DGECR-2022-00136) for making the SPARK Academy possible.
\subsubsection{\discintname}
The authors have no competing interests to declare.
\end{credits}

\bibliographystyle{splncs04}
\bibliography{references}

\end{document}